%% file: iclr2020_conference.tex
\documentclass{article} 
\usepackage{iclr2020_conference,times}

\input{math_commands.tex}

\usepackage{hyperref}
\usepackage{url}
\usepackage{graphicx}

\title{Satellite-based Prediction of Forage Conditions for Livestock in Northern Kenya}

\author{Andrew Hobbs  \\
University of California - Davis \\
\texttt{awhobbs@ucdavis.edu} \\
\And
Stacey Svetlichnaya \\
Weights \& Biases \\
\texttt{stacey@wandb.com} \\
}

\iclrfinalcopy 
\begin{document}

\maketitle

\begin{abstract}
This paper introduces the first dataset of satellite images labeled with forage quality by on-the-ground experts and provides proof of concept for applying computer vision methods to index-based drought insurance. We also present the results of a collaborative benchmark tool used to crowdsource an accurate machine learning model on the dataset. Our methods significantly outperform the existing technology for an insurance program in Northern Kenya, suggesting that a computer vision-based approach could substantially benefit pastoralists, whose exposure to droughts is severe and worsening with climate change.

\end{abstract}

\section{Introduction}

In recent years, a new literature has emerged to combine computer vision methods with satellite imagery and apply these to important challenges in developing country agriculture. One set of studies focuses on the potential to use computer vision as a less expensive and more accurate substitute for traditional forms of data collection such as crop cutting and surveys. \cite{burke2017satellite} show that satellite images are as accurate as survey data in identifying variation in agricultural yields for smallholder farmers. \cite{m2019semantic} provide the first crop type semantic segmentation dataset for smallholder farms in Africa, with labeled data from South Sudan and Ghana, and train a neural network that outperforms the previous state of the art. \cite{kamilaris2018review} review the use of convolutional neural networks (CNNs) in agricultural applications, and note that while CNNs often outperform other methods, data quality is critical to success. Our study also builds on a growing literature applying computer vision methods to aerial data more generally. \cite{maggiori2016convolutional} introduce a CNN-based framework for dense labeling of remote-sensing derived images, and \cite{mnih2012learning} find that using an appropriately chosen loss function can substantially improve results when using aerial data with noisy labels. 

Weather risk is one of the most important challenges affecting developing country farmers, and it is especially serious for the pastoralists of Northern Kenya. Droughts frequently kill large numbers of livestock, which are the primary assets owned by pastoralist households and a key source of nutrition. Worse, climate change is expected to increase drought frequency and severity \cite{seneviratne2017changes}. Index-based insurance is designed to reduce the cost of droughts by using some measurement (the index) to predict average losses in a given region. When average losses reach a threshold, index insurance products provide farmers or pastoralists with cash to offset their losses. Early index insurance products used rainfall monitors and crop cuttings to estimate average losses; more recent products have relied on remote sensing methods and satellite data. In the region studied in this paper, the International Livestock Research Institute's (ILRI) Index Based Livestock Insurance (IBLI) program has been in operation since 2011 \citep{chantarat2013designing}. It relies on an index derived from the Normalized Differenced Vegetation Index (NDVI), a remote sensing-based measure that has been shown to accurately capture changes in green vegetation.

As studied in detail by \cite{jensen2019does}, index accuracy and cost are important determinants of the value of insurance for farmers. This study explores whether a computer vision-based approach can outperform the NDVI-based approach currently used for the IBLI program in Northern Kenya. Index accuracy plays a strong role in determining the value of index insurance to farmers, and poor accuracy is a major factor limiting farmer adoption of insurance \citep{carter2017index}. Our results demonstrate that computer vision methods can generate more accurate indices for insurance purposes than existing methods and provide proof-of-concept for a machine learning-based insurance product. 

\section{Dataset}

The dataset relies on two sources: 1) ground-based labels and photos from pastoralists assessing forage conditions on site and 2) satellite images from NASA's LANDSAT mission.\footnote{In recent years, much higher resolution satellite images have become available from companies like Planet and Digital Globe. This paper uses 30m resolution data freely available from LANDSAT because higher resolution images were not available during the time period when the ground-level labels were generated. We expect that even better results will be possible using higher-resolution images in the future.} We select satellite images centered on the geolocations for which we have labeled ground-level data. This approach yields a supervised labeling of forage quality on satellite images, and it is highly generalizable. In any cases where ground-level data on an agricultural outcome is available (e.g. crop yields, disease status), this method can be used to develop a satellite-based index. Furthermore, it enables model ensembling or transfer learning across countries or indices (e.g. yields of related crops), which could improve model robustness and reduce the need for new supervised labels.

\begin{figure}[h]
\begin{center}
\includegraphics[scale=0.39]{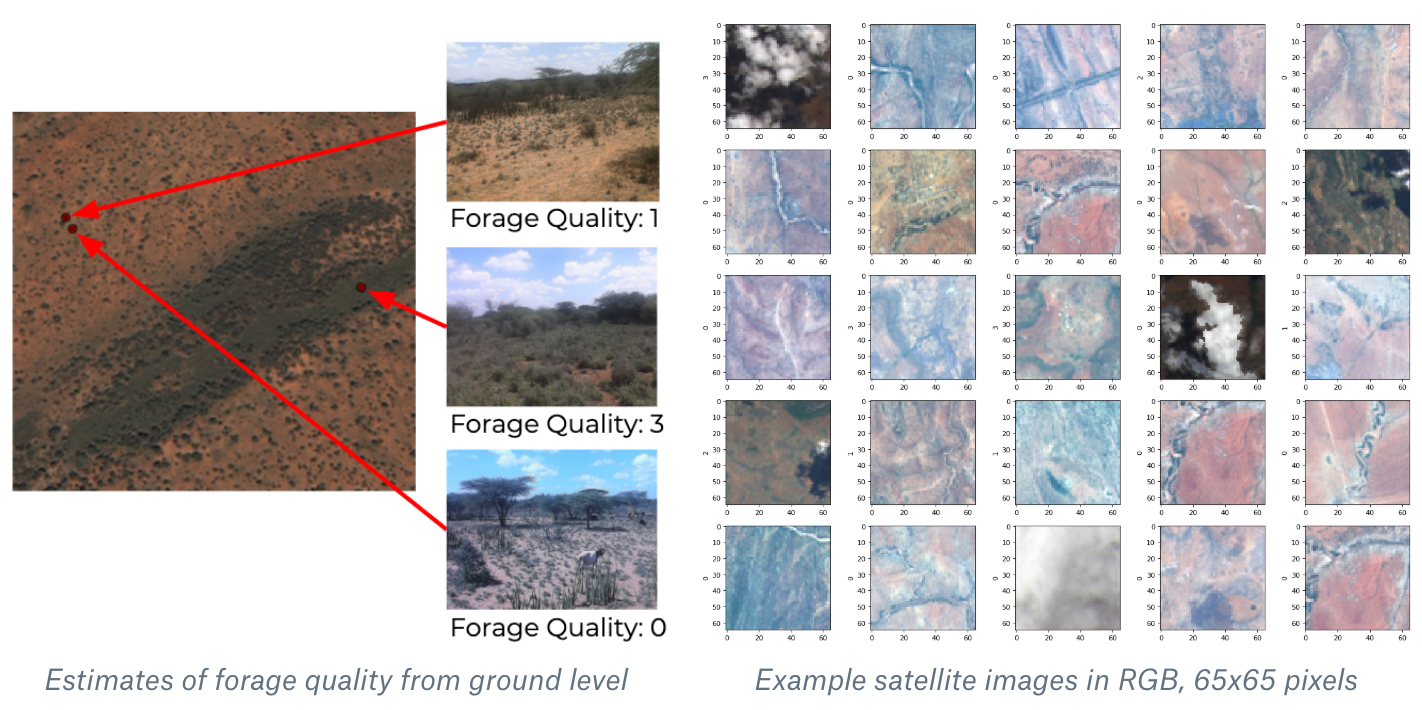}
\end{center}
\caption{Images were labeled using pastoralist report of forage conditions on the ground.}
\end{figure}

\subsection{Ground-level Labels}

A core challenge in applying computer vision methods to satellite data, particularly in developing countries, is the need for large numbers of accurate labels. Generating these labels is especially difficult because human experts cannot explicitly label aerial data with outcomes like poverty level or forage quality. A number of creative approaches have been used to circumvent this problem. \cite{sheehan2019predicting} combine georeferenced Wikipedia entries with satellite images to predict community-level asset wealth and education, and \cite{jean2016combining} and \cite{tingzon2019mapping} use nighttime lights data to train models that estimate consumption expenditure and asset wealth.

We tackle the labeling problem by leveraging a set of labels generated on the ground by pastoralist nomads--experts in assessing forage quality in the region of Northern Kenya (data collected by ILRI in 2015). At each reporting point, pastoralists were asked to estimate the number of cows that could eat for a single day within 20 meters of their location on a scale of 0-3+. Labels are discrete, so pastoralists must pick either 0, 1, 2, or 3+ based on their best estimate.

\subsection{Ground-level Images}

At each labeled location, pastoralist data collectors also took a photo of the surroundings from their vantage point. We used a subset of these photos to validate the initial pastoralist estimates through Amazon Mechanical Turk. They are included with the dataset as a potential tool for checking data quality. This verified subset of images also forms the validation set for the benchmark.

\subsection{Satellite Images}

For each labeled location, we downloaded a 65x65 pixel satellite image from within one week of the date on which the label was generated. When images were not available that met this criterion, the observation was dropped. The resulting dataset contains over 100,000 labeled images.

LANDSAT images are relatively low spatial resolution: each pixel represents a 30 meter square. However, each pixel contains more data than in typical images: in addition to the standard 3 RGB bands, LANDSAT contains 8 other bands with information outside of the visible spectrum. The near-infrared band is of particular interest, because it is known to be helpful in measuring vegetation and is an important part of the index currently used in the IBLI program.

\section{Collaborative Benchmark}

The data described above are part of an ongoing collaborative benchmark hosted by Weights \& Biases\footnote{See https://www.wandb.com/articles/droughtwatch to learn more about the benchmark}. The platform differs from many others in the sense that all participants can share their code and training workflows, discuss their approaches, and potentially build on each others' findings. This collaborative structure is designed to promote faster progress and knowledge sharing across participants \citep{wandb}.

\begin{figure}[h]
\begin{center}
\includegraphics[scale=0.45]{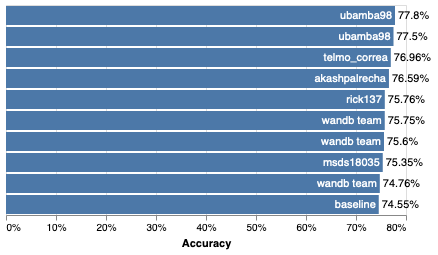}
\end{center}
\caption{The top ten entries as of February 16, 2020.}
\end{figure}

The highest accuracy reported at the time of writing is 77.8\%. The model is based on the EfficientNet architecture proposed by \cite{tan2019efficientnet}.\footnote{Details on this as well as other leading models can be examined at https://app.wandb.ai/wandb/droughtwatch/benchmark/leaderboard} This model slightly outperformed several ResNet-based models and a tuned version of the CNN baseline. The CNN baseline is an intentionally simple example for the benchmark: three convolutional blocks with max pooling, dropout on blocks 2 and 3, and two fully-connected layers before the softmax.  The layer sizes and hyperparameters are easily configurable to encourage participants to explore model variants. All of the models in the leaderboad significantly outperform an NDVI-based model, which achieved a predictive accuracy of just 67\%. 

Since the difference between a basic CNN and a ResNet is only a few percent accuracy, there are clear opportunities for improvement: tuning different network architectures, loss functions, optimizers, and other hyperparameters. On the data side, promising next steps include a finer-grain analysis of the spectral bands (of the 11, a few seem to add more noise than signal, such that removing them improves accuracy), filtering out images with obscuring clouds, data augmentation (e.g. rotate, flip), and addressing the class imbalance (roughly 60\% of the data is of class 0, classes 1 and 2 have 15\% each, and the remaining 10\% is class 3+). The benchmark collaboration will continue to support and cross-pollinate these model improvements.

\section{Analysis}

In this section we explore several extensions to the modeling. First, we explore how much predictive accuracy is lost when the images are center-cropped, reducing context. There is little loss of accuracy even when cropping to 35x35 pixel rather than 65x65 pixel images. Second, we explore reframing the problem as an ordinal regression problem to ensure that predictions are close even if not perfect.

\subsection{Image Size}
Because each pixel is 30 meters across, the 65x65 pixel images in the dataset represent nearly 1 kilometer squares. While spatial context likely helps generate better predictions, it is intuitive that these images may be larger than necessary. Sensitivity tests indicate that networks trained on much smaller squares are able to perform similarly well: prediction accuracy did not decrease substantially in our tests until images were smaller than 35x35. Smaller images also reduce training and prediction time (since new image data must be downloaded from a server).

\subsection{Ordinality}

The benchmark and initial analysis relied on standard neural network architecture for categorical outputs: labels encoded as one-hot vectors and the final layer as a softmax activating at most one neuron in the output layer. A prediction is accurate only if it is exactly correct, which means that mistakenly categorizing a `3' as a `0' is scored identically to categorizing a `2' as a `1', even though the second is intuitively a smaller error.

Ordinality is especially important for this problem for two reasons. First, there is inherent subjectivity and uncertainty in the labeling process. Two expert pastoralists could look at the same 20 meter circle, and one might estimate it could support one cow while the other would estimate two. It's much less likely that one would say `3+' and the other `0.' Second, for effective index insurance it is not essential that we have an exactly correct estimate of conditions at each point in space and time, but it \textit{is} essential that our index only make small errors. While a small error might lead the insurance product to slightly underpay in the event of a drought, a large error might lead to a complete failure to pay, defeating the purpose of the product.

We follow the approach described by \cite{cheng2008neural}, replacing the one-hot vectors with a format that preserves information on ordinality within the sample. For example, a location that can support 0 cattle is encoded [0,0,0], 1 is encoded as [1,0,0], 2 as [1,1,0], and 3+ as [1,1,1]. Each entry in the output vector can be thought of as predicting whether the carrying capacity is greater than or equal to some value: a `1' in the first position of the array means that the location can support one or more animals; a `1' in the second position means that it can support two or more, etc. Rather than a softmax, the final layer uses a sigmoid activation function to activate more than one output. 

Ordinal accuracy can be evaluated with a binary cross-entropy loss function and binary accuracy to calculate the share of matching outputs between the prediction and target. In other words, a `2' categorized as a `1' receives an accuracy of 66.66\%, while a `3' categorized as a `0' receives an accuracy of 0\% . Scored this way, our ordinal model achieves a validation accuracy of 87.88\%. Because of the difference in scoring and loss functions, the accuracy from the ordinal regressions is not directly comparable to the categorical models. However, evaluating the best categorical model with ordinal accuracy would only improve the metric: some of the errors previously counted as 0\% would add 33.33\% or 66.66\%, depending on the distance between the labels.

\section{Conclusion}

Computer vision methods have great promise as a means of shielding farmers from the negative effects of climate change due to their ability to flexibly map satellite data to information on yields, forage quality, or other measures collected from the ground. This paper provides an initial dataset for exploring the potential of computer vision for index insurance and proof of concept that computer vision methods can outperform existing remote sensing techniques. Once trained and deployed, computer vision models can assess conditions on demand, making it possible to monitor more frequently and compensate the insured more quickly. By increasing both the accuracy and the speed of insurance payments, computer vision has the potential to substantially increase the benefit of insurance for farmers.


\subsubsection*{Acknowledgments}

We would like to thank Nathan Jensen at the International Livestock Research Institute for making the forage quality data we describe in this paper available for the benchmark competition we describe in this paper.

The forage quality data used in this paper was collected through a research collaboration between the International Livestock Research Institute, Cornell University, and UC San Diego. It was supported by the Atkinson Centre for a Sustainable Future’s Academic Venture Fund, Australian Aid through the AusAID Development Research Awards Scheme Agreement No. 66138, the National Science Foundation (0832782, 1059284, 1522054), and ARO grant W911-NF-14-1-0498.

\bibliography{iclr2020_conference}
\bibliographystyle{iclr2020_conference}


\end{document}

%% file: math_commands.tex
\usepackage{amsmath,amsfonts,bm}

\def\eqref#1{equation~\ref{#1}}

\def\1{\bm{1}}

\DeclareMathAlphabet{\mathsfit}{\encodingdefault}{\sfdefault}{m}{sl}
\SetMathAlphabet{\mathsfit}{bold}{\encodingdefault}{\sfdefault}{bx}{n}

